%% file: main.tex
\documentclass[10pt,twocolumn,letterpaper]{article}

\usepackage[pagenumbers]{cvpr} 


\definecolor{cvprblue}{rgb}{0.21,0.49,0.74}
\usepackage[pagebackref,breaklinks,colorlinks,allcolors=cvprblue]{hyperref}
\usepackage{xcolor, colortbl, booktabs, multirow, subcaption, dashrule, tikz, algorithm, algpseudocode}
\usepackage[accsupp]{axessibility}  
\definecolor{mygreen}{RGB}{144,238,144} 

\def\paperID{6} 
\def\confName{CVPR}
\def\confYear{2025}

\title{Wheat3DGS: In-field 3D Reconstruction, Instance Segmentation and Phenotyping of Wheat Heads with Gaussian Splatting}

\author{Daiwei Zhang$^{1*}$ \quad Joaquin Gajardo$^{1*\dagger}$ \quad Tomislav Medic$^1$ \quad Isinsu Katircioglu$^2$ \\ \quad Mike Boss$^1$ \quad Norbert Kirchgessner$^1$ \quad Achim Walter$^1$ \quad Lukas Roth$^1$ \vspace{0.3em} \\
{\normalsize $^1$ETH Zürich} \quad
{\normalsize $^2$Swiss Data Science Center} \quad
{\small $^*$Equal contribution} \quad
{\small $^\dagger$Corresponding author: \texttt{jgajardo@ethz.ch}} \\
{\small Webpage: \texttt{\url{https://zdwww.github.io/wheat3dgs/}}} \\
}

\begin{document}
\maketitle


\input{sec/0_abstract}    
\input{sec/1_intro}
\input{sec/2_related}
\input{sec/3_method}

\input{sec/4_data}

\input{sec/5_results}

\input{sec/6_discussion}
\input{sec/7_conclusion}
{
    \small
    \bibliographystyle{ieeenat_fullname}
    \bibliography{main}
} 

\input{sec/X_suppl}

\end{document}

%% file: sec/0_abstract.tex
\begin{abstract}
Automated extraction of plant morphological traits is crucial for supporting crop breeding and agricultural management through high-throughput field phenotyping (HTFP).
Solutions based on multi-view RGB images are attractive due to their scalability and affordability, enabling volumetric measurements that 2D approaches cannot directly capture.
While advanced methods like Neural Radiance Fields (NeRFs) have shown promise, their application has been limited to counting or extracting traits from only a few plants or organs.
Furthermore, accurately measuring complex structures like individual wheat heads—essential for studying crop yields—remains particularly challenging due to occlusions and the dense arrangement of crop canopies in field conditions.
The recent development of 3D Gaussian Splatting (3DGS) offers a promising alternative for HTFP due to its high-quality reconstructions and explicit point-based representation.
In this paper, we present Wheat3DGS, a novel approach that leverages 3DGS and the Segment Anything Model (SAM) for precise 3D instance segmentation and morphological measurement of hundreds of wheat heads automatically, representing the first application of 3DGS to HTFP.
We validate the accuracy of wheat head extraction against high-resolution laser scan data, obtaining per-instance mean absolute percentage errors of 15.1\%, 18.3\%, and 40.2\% for length, width, and volume. We provide additional comparisons to NeRF-based approaches and traditional Muti-View Stereo (MVS), demonstrating superior results. 
Our approach enables rapid, non-destructive measurements of key yield-related traits at scale, with significant implications for accelerating crop breeding and improving our understanding of wheat development.

\end{abstract}
\vspace*{-3mm}

%% file: sec/1_intro.tex
\begin{figure}[htbp]
    \centering
    \includegraphics[width=\linewidth]{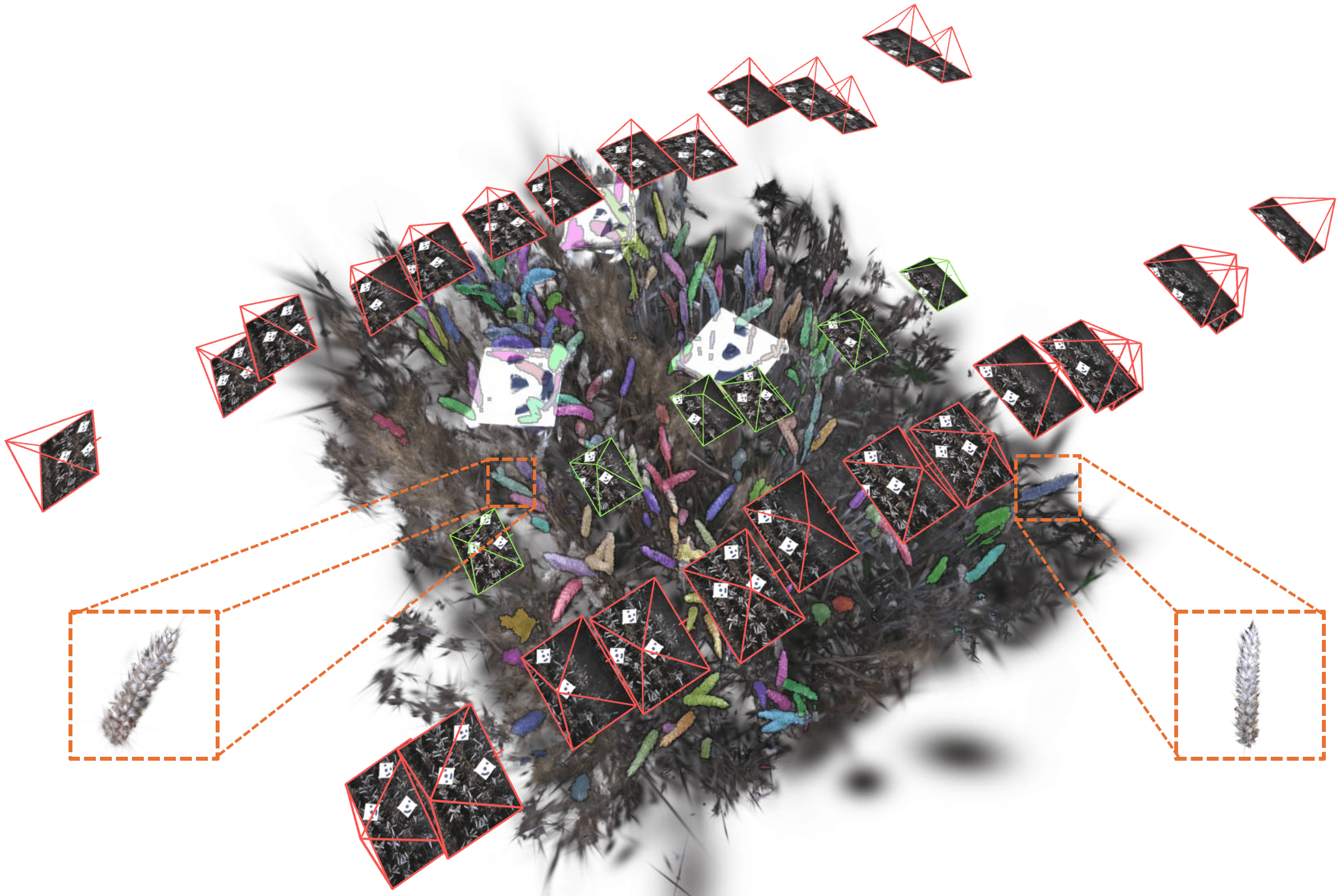}
    \caption{3D Gaussian Splatting reconstruction of a wheat plot with segmented 3D wheat heads instances (in different colors). We use 30 views for reconstruction (red frustums) and 6 out-of-distribution views for evaluation (green frustums).}
    \label{fig:title}
    \vspace*{-4mm}
\end{figure}

\section{Introduction}
\label{sec:intro}

Accurate and rapid measurement of plant traits is essential for advancing crop breeding programs and understanding plant development~\cite{Araus18a}. In particular, the precise characterization of wheat head morphology—including length, width, and volume—is critical for assessing yield potential and phenotypic variation of this staple crop~\cite{Hund2019}.  Plant phenotyping traditionally relies on laborious manual measurements. Capturing and analyzing plant point clouds obtained with laser scans~\cite{liu2023extraction,wang2022unsupervised, medic2023towards} and Multi-View Stereo (MVS)~\cite{Klodt14a,Nguyen16a,Wu20a,Huang24a,Cui25a} provide automated alternatives, but are either too costly, slow or lack the fine-grained details required for accurate morphological measurements.
Implicit neural representations, such as Neural Radiance Fields (NeRFs)~\cite{Mildenhall22a,Mueller22a,Tancik23a}, have emerged as a promising alternative for image-based plant phenotyping by overcoming limitations of traditional MVS approaches. By modeling space continuously, neural representations enable a more detailed reconstruction of intricate plant structures~\cite{Arshad24a} and are better equipped to represent occluded regions and recover fine-grained details. Despite these advantages NeRFs remain computationally expensive, and require dense sampling of a neural network to obtain point clouds ~\cite{yu_and_fridovichkeil2021plenoxels}, making editing and post-processing cumbersome. 

While NeRFs have been applied in field conditions, their use has been limited to small-scale studies as proof-of-concept approaches or to measure simple traits on a handful of plants~\cite{Saeed23a,Yang24a}.
Recent advancements in point-based 3D representations, particularly 3D Gaussian Splatting (3DGS)~\cite{Kerbl23a}, offer a more efficient alternative of radiance fields for 3D reconstruction. 3DGS directly represents the scene with explicit Gaussian primitives, enabling fast rasterization and a more straightforward extraction of geometric features without additional post-processing. Although 3DGS has recently begun to be explored for plant phenotyping, its applications have so far been limited to controlled indoor environments at the single-plant level~\cite{Ojo24a, Shen24a}. Meanwhile, high-throughput field phenotyping (HTFP) is essential to support crop breeding programs and for yield prediction, but significant challenges arise in outdoor field conditions, such as heavy occlusions caused by other leaves or wheat heads in dense canopies. Even simple but critical tasks such as accurately detecting and counting wheat heads remain challenging \cite{david2105global}, and common practice is still to do this manually.

To address these challenges and push the current capabilities, we propose Wheat3DGS, a novel pipeline that leverages 3DGS and the Segment Anything Model (SAM) \cite{kirillov2023segment} for 3D reconstruction of outdoor wheat canopies and 3D wheat head instance segmentation, enabling their individual extraction and measurement (\autoref{fig:title}). To address the challenge of segmenting individual wheat heads, we obtain per-view segmentation masks by prompting SAM with bounding boxes provided by an off-the-shelf wheat head detector---the winning model\footnote{\url{https://github.com/ksnxr/GWC_solution}} of the 2021 Global Wheat Head Detection Challenge~\cite{david2105global}. These masks are associated in 3D by projecting them onto the Gaussian splats, enabling the annotation and extraction of individual wheat heads in 3D space as groups of Gaussians. This hybrid approach combines the efficiency of explicit 3D representations with the semantic precision of advanced 2D vision models, facilitating accurate organ-level trait extraction.

We validate our method through comprehensive evaluation of canopy reconstruction quality, wheat head segmentation, and trait measurement quality.
Our results demonstrate that Wheat3DGS outperforms NeRF-based in canopy reconstruction and 3D segmentation abilities, and exceeds MVS in wheat head trait measurement quality.
Our main contributions can be summarized as follows:

\begin{itemize}
\item We show that 3D Gaussian Splatting can be effectively used for creating detailed 3D reconstructions of crop canopies from overhead RGB imagery and provide a quantitative and qualitative comparison to NeRF-based methods.

\item We propose a method to perform 3D wheat head instance segmentation on a 3D reconstructed canopy by leveraging a pretrained wheat head detector and SAM, and associating semantic information in 3D. We also provide quantitative evaluation for extracted traits, and compare against high-resolution laser scan data, demonstrating superior accuracy and efficiency compared to MVS.

\item We release a dataset comprising RGB images with calibrated camera poses, corresponding laser scans for seven wheat plots, and the view-consistent segmentation masks generated by our approach, providing a valuable resource for future research in image-based plant phenotyping.
\end{itemize}

By combining state-of-the-art 3D reconstruction methods with advanced segmentation techniques, Wheat3DGS addresses key challenges in automated plant phenotyping, improving our ability to measure wheat head morphology, and lays the foundation for future large-scale studies on crop development and yield prediction.

%% file: sec/2_related.tex
\section{Related Work}

\paragraph{3D reconstruction.} Traditional 3D reconstruction methods such as Structure-from-Motion (SfM)~\cite{Triggs00a,Snavely06a,Schonberger16a} and MVS~\cite{Seitz06a,Goesele07a,Yao20a,Wei21a,Zhang23a} estimate scene geometry by matching keypoints and triangulating points across multiple images. However, they require numerous images, precise matching, and high computational power, while struggling with occlusions, textureless regions, and scalability.

With the rise of neural rendering \cite{tewari2022_advances}, data-driven techniques have transformed the field by enabling high-fidelity 3D reconstruction and novel view synthesis (NVS) with significantly fewer input images. Notably, NeRFs~\cite{Mildenhall22a} introduced an implicit scene representation with coordinate-based Multi Layer Perceptrons (MLP) that can produce highly realistic renderings, but require large training times due to an expensive volumetric rendering process. More recently, 3DGS~\cite{Kerbl23a} has emerged as an efficient alternative, adopting an explicit representation with anisotropic 3D Gaussian primitives and tiled rasterization to achieve real-time rendering while maintaining high visual quality. These advancements mark a paradigm shift in 3D reconstruction, bridging the gap between traditional geometry-based methods and neural approaches. Building on 3DGS, \cite{Guedon23a} introduces an algorithm for mesh extraction and a regularization term to encourage 3D Gaussians to align with a surface and facilitate mesh extraction, while \cite{Huang24a} simplifies 3D modeling by adopting flat 2D Gaussians, enabling faster rendering and reduced storage requirements. However, 3D and 2DGS models focus solely on scene appearance and geometry, lacking object-level understanding.~\cite{Ye23a} addresses this by lifting 2D semantic masks to 3D with identity-encoded Gaussians for instance grouping. Yet, their method relies on view-consistent masks obtained by a video object tracker, making it prone to failures with similar or intermittently occluded objects. In contrast,~\cite{Lyu24a} employs 3D-aware mask association, matching projected Gaussians to 2D masks and assigning group IDs based on maximum overlap, leading to a better differentiation of similar objects. Similarly, ~\cite{Shen24a} introduces an optimal solver for enhancing the accuracy and efficiency of embedding 2D semantic masks in 3DGS reconstructions.

\paragraph{Plant reconstruction.} Capturing 3D information for plant phenotyping has traditionally relied on ranging sensors like LiDAR \cite{Kronenberg20a}, RGB-D cameras like Intel RealSense~\cite{Pan23a}, or RGB-based Structure-from-motion (SfM) and MVS~\cite{Murakami12,Bendig13a,Gillan14,Roth18a,Sunvittayakul22a,Drofova23a}.  However, these methods are often costly and struggle to capture thin plant structures, leading to noisy and sparse point clouds. To address these challenges,~\cite{Esser23a} uses a robotic platform equipped with both LiDAR and camera sensors to reconstruct 3D plant structures from multiple sensing modalities. More recently, methods relying entirely on RGB images have emerged. For instance, \cite{Arshad24a,Hu24a,Zhai24a,Zhang24a} evaluate several NeRF variants~\cite{Chen22a,Mueller22a,Tancik23a} for 3D reconstruction and NVS of various plants, including corn, tomatoes, and fruit trees, across different levels of complexity. However, these methods focus solely on 3D reconstruction without explicit plant phenotyping. To integrate plant trait analysis, PeanutNeRF~\cite{Saeed23a} employs Nerfacto~\cite{Tancik23a}, a fast NeRF variant based on \cite{Mueller22a}, for both 3D reconstruction and phenotypic analysis, extracting traits such as node count and flowering. However, it is limited to coarse plant structures and applies only to isolated plants in controlled environments with minimal occlusion. Similarly,~\cite{Yang24a} relies on NeRF for 3D reconstruction of rice panicles, combining YOLOv8 and SAM for instance segmentation and trait estimation, such as length and volume. However, their approach focuses on reconstructing single rice panicles one at a time, limiting its applicability in real-word scenarios. More scalable solutions are proposed by ~\cite{Smitt24a,Meyer24a}, which extend NeRFs by mapping a 3D point not only to density and color but also to semantic information, successfully segmenting dozens of fruits in orchards and greenhouses.

3DGS has been proposed as a promising alternative to NeRF-based plant phenotyping~\cite{Ojo24a,shen25a_plantgaussian,Stuart25_high-fidelity}. However, these studies have so far remained exploratory at the single plant level, without providing semantic insights for plant trait analysis. In this work, we propose a mechanism to identify and segment wheat head instances in 3DGS reconstructions of wheat canopies in field conditions, thus representing the first work using radiance fields for 3D phenotypic trait extraction at scale.

%% file: sec/3_method.tex
\section{Methodology}
\begin{figure*}[ht]
    \centering
    \begin{minipage}{0.99\textwidth}
        \centering
        \includegraphics[width=\textwidth]{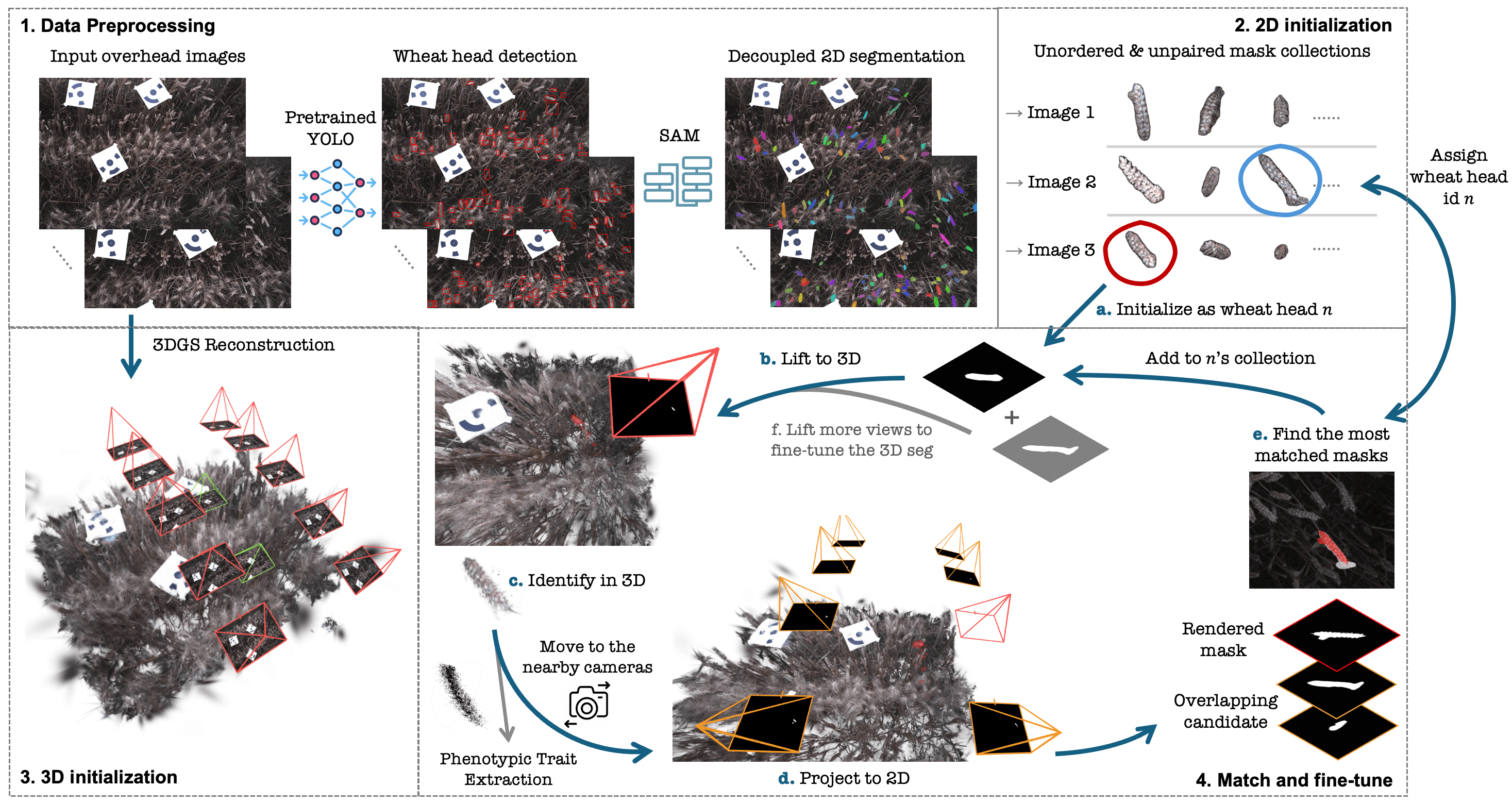}
    \end{minipage}\hfill
    \caption{\textbf{Overview of our pipeline}: Given a set of RGB images capturing our target wheat field plot with a system of overhead cameras, we extract 2D segmentation masks of detected wheat heads (\autoref{sec:segmentation_masks}) and reconstruct a 3D representation of the plot using 3D Gaussian Splatting  (\autoref{sec:3D_reconstruction}) as initialization. For robust 3D wheat head segmentation (\autoref{sec:3D_wheat_segmentation}), we propose a \textbf{match-and-fine-tune} strategy (\autoref{sec:multi-view-asociation}) that iteratively associates collections of decoupled masks and refines the 3D Gaussian representation of each segmented wheat head by alternating between lifting 2D masks to 3D and projecting 3D segmentations back to other views.}
    \label{fig:pipeline}
    \vspace*{-5mm}
\end{figure*}

Our method (\autoref{fig:pipeline}) is divided into four main sub-parts: wheat head segmentation on 2D images (\autoref{sec:segmentation_masks}), 3D reconstruction of wheat canopies (\autoref{sec:3D_reconstruction}), 3D instance segmentation of individual wheat heads (\autoref{sec:3D_wheat_segmentation} and \autoref{sec:multi-view-asociation}), and phenotypic trait extraction (\autoref{sec:trait_extraction}).

\subsection{2D Wheat Head Segmentation}
\label{sec:segmentation_masks}
We start with a collection of $\mathcal{C}$ unposed multi-view input images $\mathcal{I} = \{ I_1, \dots, I_C\}$ of a wheat field. Detecting and precisely segmenting individual wheat heads from real-world images remains challenging due to their randomly scattered and densely-packed distribution. To address this, we adopt a two-step process that relies on a pre-trained YOLOv5 model \cite{redmon2016you} fine-tuned on the Global Wheat Head Dataset \cite{david2105global}, which covers different locations, development stages, and capture conditions, followed by segmentation with SAM \cite{kirillov2023segment}. We utilize the pretrained YOLOv5 model given its publicly available weights specifically optimized for wheat head detection, and to demonstrate the robustness of our method.
We use this model to generate a set of bounding boxes for each image $I_i$, $\mathcal{B}^i = \{b_1, \dots, b_{N_i} \}$, where $N_i$ is the number of detected wheat heads on it, and provide them as prompts to SAM~\cite{kirillov2023segment} individually, to obtain precise 2D segmentation masks for each detected wheat head. By iterating this procedure over all detections and images, we obtain a collection of 2D segmentation masks $\mathcal{M} = \{\mathcal{M}^1, \dots ,\mathcal{M}^C\}$, where $\mathcal{M}^i = \{ M_1, \dots, M_{N_i} \}$ is the set of single-wheat-head semantic masks associated with image $I_i$. Since SAM's segmentation output is generated independently for each image and is instance-agnostic, two distinct masks $M$ from different images can correspond to the same physical wheat head in the canopy. 




\subsection{3D Gaussian Splatting}
\label{sec:3D_reconstruction}

We adopt 3DGS \cite{Kerbl23a} as our scene representation for 3D reconstruction given its more straightforward editing capabilities compared to NeRF-based representations. Specifically, a 3D scene---one wheat plot in our case, as described in \autoref{sec:data}---is parameterized as a set of learnable 3D Gaussian primitives $\mathcal{G} = \{ G_k \}_{k=1}^K$. Each Gaussian $G_k$ is characterized by its centroid position $\mathbf{p}_k \in \mathbb{R}^3$, 3D scale vector $\mathbf{s}_k \in \mathbb{R}^3$, a quaternion $\mathbf{q}_k \in \mathbb{R}^4$ representing rotation, opacity $\alpha_k \in \mathbb{R}$, and color features $\mathbf{c}_k$ encoded by spherical harmonics (SH) coefficients. In its rendering process, 3DGS adopts a point-based rasterization approach that blends 3D Gaussians sorted by depth onto a 2D image plane using alpha compositing. A pixel-feature $X$ is computed as:
\begin{equation}
    X = \sum_{k \in \mathcal{K}} x_k \alpha_k \prod_{j=1}^{k-1} (1-\alpha_j) = \sum_{k \in \mathcal{K}} x_k a_k T_k
     \label{eq:gs}
\end{equation}
\noindent where the property $x_k$ can be view-dependent color, depth, or other optimizable features of each Gaussian $G_k$, and $T_k$ is the transmittance. The rendered image is compared to the ground truth camera view via photometric loss, which provides the learning signal to update the parameters of the Gaussians involved in rendering each pixel of the image.

\subsection{3D Wheat Head Segmentation}
\label{sec:3D_wheat_segmentation}
In this section we provide the preliminaries and the problem definition for 3D segmentation.
Given a reconstructed 3DGS scene of a wheat field plot parameterized by 3D Gaussians $\mathcal{G}$ and containing a set of wheat heads $\mathcal{N}$, our goal is to identify the subset $\mathcal{G}_n \subset \mathcal{G}$ for each distinct wheat head $n \in \mathcal{N}$. 

Suppose we have $\mathcal{L}$ 2D binary masks exclusively associated with one wheat head $n$, denoted as $\{ M^{\ell} \}_n$ where $\ell \in \mathcal{L}$, $\lvert \mathcal{L} \rvert \leq \lvert \mathcal{C} \rvert$, and pixels with value 0 represent the background and 1 denotes the foreground (i.e. wheat head). This naturally leads to a 3D scene segmentation problem: assign a binary label $W_k \in \{ 0, 1\}$ to each 3D Gaussian $G_k$, indicating whether it corresponds to the targeted wheat head, by projecting the 2D binary masks $M^{\ell}$ into the 3D space.

In contrast to \cite{ye_gaussian_grouping_2023} that uses gradient descent to iteratively optimize a learnable embedding (from which the label assignment $W_k$ can be derived) as a feature in \autoref{eq:gs}, we adopt the method introduced in \cite{Shen24a}, which directly solves for the label $W_k$ in closed form via integer linear programming. 
For each Gaussian $G_k$ in a reconstructed scene $\mathcal{G}$, we set $x_k=W_k$ in \autoref{eq:gs} and optimize $W_k$ while keeping all other attributes fixed. The 3D segmentation problem for a specific wheat head can then be formulated as solving for $\{ W_k \}$ by minimizing the objective function $\mathcal{F}$ defined as the mean absolute error between the rendered 2D segmentation mask and the ground truth 2D mask:
\begin{equation}
    \begin{aligned}
        \min_{\{ W_k \}} \quad \quad \mathcal{F} = &\sum_{\ell \in \mathcal{L}}  \left| \mathcal{R} \left( \{ G_k\}, \{ W_k \}\right)- M^{\ell} \right| \\
        = &\sum_{\ell \in \mathcal{L}} \sum_{p \in M^{\ell}} 
        \left| \sum_{k \in \mathcal{K}} W_k \alpha_k T_k - M^{\ell}(p) \right|
    \end{aligned}
    \label{eq:ILP}
\end{equation}
\noindent subject to $W_k \in \{ 0, 1 \}$, where $\mathcal{R}$ is the differentiable rasterizer that renders each pixel value by blending $\mathcal{K}$ depth-sorted Gaussians, $p$ represents a pixel in the provided binary mask $M^{\ell}$, and $\ell \in \mathcal{L}$ specifies the view from which $M^{\ell}$ is available. We follow the approach in \cite{Shen24a} to solve the optimal assignment for $\{ W_k \}$ using  majority vote across views and a background bias to account for noise in the masks. 

Intuitively, the more ground truth masks $M^{\ell}_n$ are provided from different views $\ell$ for a wheat head $n$, the more accurately the subset of 3D Gaussians $\mathcal{G}_n = \{ G_k \mid W_k = 1\}$ will represent the actual wheat head.

\subsection{Multi-view Instance Association}  
\label{sec:multi-view-asociation}

The main challenge in our problem setup is the absence of associations between 2D wheat head masks detected across different views, i.e. $\{ M^{\ell} \}_n$, as defined in \autoref{sec:3D_wheat_segmentation}, is unknown.

Existing methods \cite{ye_gaussian_grouping_2023} either employ a video tracker \cite{cheng2023tracking} to propagate and associate 2D masks, which is ineffective in our case due to sparse viewpoints, and the densely-packed and repetitive structure of wheat canopies, or require hand-crafted point prompts \cite{Shen24a}. Thus, we developed a fully automatic iterative \textbf{match-and-fine-tune} strategy to address this challenge effectively for all wheat heads.

For a specific wheat head $n$, we are only certain that a single binary mask $M^{\ell}_n$ from one view $\ell$ is associated with it. By minimizing the discrepancy between the rendered ($\hat{M}^{\ell}_n$) and the provided mask ($M^{\ell}_n$), we can optimize the binary label $W_k \in \{ 0, 1\}$ of each 3D Gaussian $k$ as outlined in \autoref{sec:3D_wheat_segmentation}. However, intuitively, since only a single view $\ell$ is available, the estimated set $\{ \hat{W}_k \}$ will be less accurate in representing the complete 3D structure of the wheat head when lifting the 2D segmentation to 3D.

To improve accuracy, we project the estimated $\{ \hat{W}_k \}$ to another view $\ell^{\prime} \in \mathcal{L}$, where $\ell^{\prime} \neq \ell$, and render a binary mask $\hat{M}^{\ell'}_n$. Such rendered masks are often distorted due to insufficient views for accurate 3D segmentation.
Hence, for each projected mask $\hat{M}^{\ell'}_n$ in a camera view, we identify the potential matching binary mask $M^{\ell^\prime}_{n^\prime}$ with the highest Intersection over Union (IoU) from the previously obtained 2D segmentation masks collection, which corresponds to a candidate matching wheat head $n^{\prime}$. If the precision between $\hat{M}^{\ell'}_n$ and $M^{\ell'}_{n^\prime}$ (we use precision because we observe $\hat{M}$ is often stretched) is larger than an empirically set threshold of 0.8, then we propose $n=n^\prime$, that is, $M_n^\ell$ and $M^{\ell'}_{n^\prime}$ correspond to the same wheat head. We continue this procedure for the remaining views in $\mathcal{L}$ and collect a new set of binary masks $\mathcal{M}_n= \{ M^{\ell}_{n}, M^{{\ell}^\prime_1}_{n}, M^{{\ell}^\prime_2}_{n}, \dots\}$, representing the same matched wheat head $n$ across different views. Note that we often have $\lvert \mathcal{M}_n \rvert < \lvert \mathcal{L} \rvert$ due to the limited camera coverage and the missed detection of wheat heads in 2D. We again solve for the linear optimization in \autoref{eq:ILP}, but now restrict the summation to include only masks $M^\ell \in \mathcal{M}_n$. A weighted majority vote approach, as introduced in \cite{Shen24a}, is used to resolve the contradiction within the mask set. The output assignment $\{ \hat{W}_k \}$ now better identifies the 3D Gaussians belonging to wheat head $n$, that is, we have found a subset $\mathcal{G}_n = \{ G_k \mid \hat{W}_k = 1\}$ that more accurately represents wheat head $n$ from its matching detections across views. We then assign a unique wheat head ID $n>0$ as an additional attribute to each Gaussian $G_k \in \mathcal{G}_n$. Finally, we exclude the matched masks $\mathcal{M}_n$ from the collection for further iterations, and repeat the process until there are no masks left to process.






\subsection{3D Phenotypic Trait Extraction}
\label{sec:trait_extraction}

Once 3D point clouds of individual wheat head instances were obtained, they were subject to preprocessing and trait extraction steps. The preprocessing step was realized as follows: 1) random subsampling to 5000 points (if greater than 5000); 2) running HDBSCAN \cite{mcinnes2017hdbscan} to extract the dominant cluster of points---likely to correspond to a wheat head; 3) running robust Statistical Outlier Removal (SOR). 
Subsequently, we obtained per wheat head length, width, and volume. For each wheat head, length is extracted by projecting 3D points onto a plane spanning through the first and second principal components (1st-2nd-PC plane), fitting a 2D smoothing spline, and evaluating an approximation of the related arc length integral. Width is computed as the robust maximum distance (99th percentile) of points from the 1st-2nd-PC plane, and volume is computed from the convex hull obtained with the Quickhull algorithm \cite{Barber96a}. All (hyper)parameters were chosen by trial and error to ensure generalizability across the used datasets. 
Further implementation details are given in the accompanying open-source code. The extracted traits are analyzed in \autoref{sec:results-fruit_sizing}.


%% file: sec/4_data.tex
\section{Data}
\label{sec:data}

\paragraph{Setup.}
Data collection was performed on July 17 2024, and relied on a small-scale wheat phenotyping experiment. The setup comprised seven plots, each measuring approx. 1.5 m\textsuperscript{2} and having six seeding rows, each row related to a different wheat genotype (see \autoref{fig:title} for a plot example).
A setup overview is presented in the suppl. material.

\paragraph{Images.}
Image acquisition was conducted using the cable-mounted camera rig system from ETH Zürich's Field Phenotyping Platform (FIP) \cite{Kirchgessner2017}. We captured 36 images (12 MP) per plot from 12 identical cameras with 35 mm lenses. Three coded markers placed on each plot facilitated SfM, scale setting, and alignment with reference scans. SfM was performed in Agisoft Metashape (St. Petersburg, Russia) to obtain camera calibrations and sparse point clouds. Additionally, we generated dense MVS point clouds for comparison with our proposed workflow. 

\paragraph{Laser scans.}
 Reference measurements were obtained on the same day using a FARO Focus 3D S 120 (FARO Technologies, Inc, FL, USA) terrestrial laser scanner (TLS) with full resolution (1.6 mm @ 10 m) and the highest quality setting. The scanning setup comprised 19 scans at a few meters distance, aside and above the canopy using a tripod and a custom mount (see suppl. material). The scans were registered in the FARO SCENE 2022.1.0 software using a target-based algorithm, relying on six laser scanning reference spheres placed within the scene, and achieving a mean alignment error of 3~mm. This was followed by a coarse alignment with the dense MVS point cloud using the Kabsch algorithm \cite{kabsch1976solution} and corresponding marker points identified in both point clouds. Finally, the registered scans were precisely aligned with 3DGS centroids of all wheat heads (obtained following \autoref{sec:multi-view-asociation}) using the ICP algorithm \cite{besl1992method} on the subsampled point clouds. The alignment of the corresponding scene elements between the TLS and 3DGS datasets was assessed to be within 10 mm on average based on visual inspection. As scanning took approximately 6~h, the internal geometry of the scene could change during the acquisition due to mild wind gusts and plant motion, which affects the quality of the scan registration \cite{medic2023challenges}. 
 Hence, even though TLS is commonly considered as ``ground truth'' for built and urban environments, in this challenging scenario, it should be considered as an independent control of comparable reconstruction quality.

%% file: sec/5_results.tex
\section{Results}
In this section, we present results in terms of NVS (\autoref{sec:results-nvs}), wheat head detection and segmentation by comparing to state-of-the-art models (\autoref{sec:results-seg}), and geometric validation against TLS data (\autoref{sec:results-fruit_sizing}).


\subsection{Novel view synthesis} \label{sec:results-nvs}

We evaluated different differentiable rendering methods on our dataset to determine the optimal underlying scene representation for 3D segmentation. In order to robustly assess the reconstruction quality, we used 30 images for training and withheld 6 images for evaluation. For all plots, we selected the test views such that they were out-of-the-distribution compared to the train views (see \autoref{fig:title}). 

\begin{figure}
    \centering
    \includegraphics[width=\columnwidth]{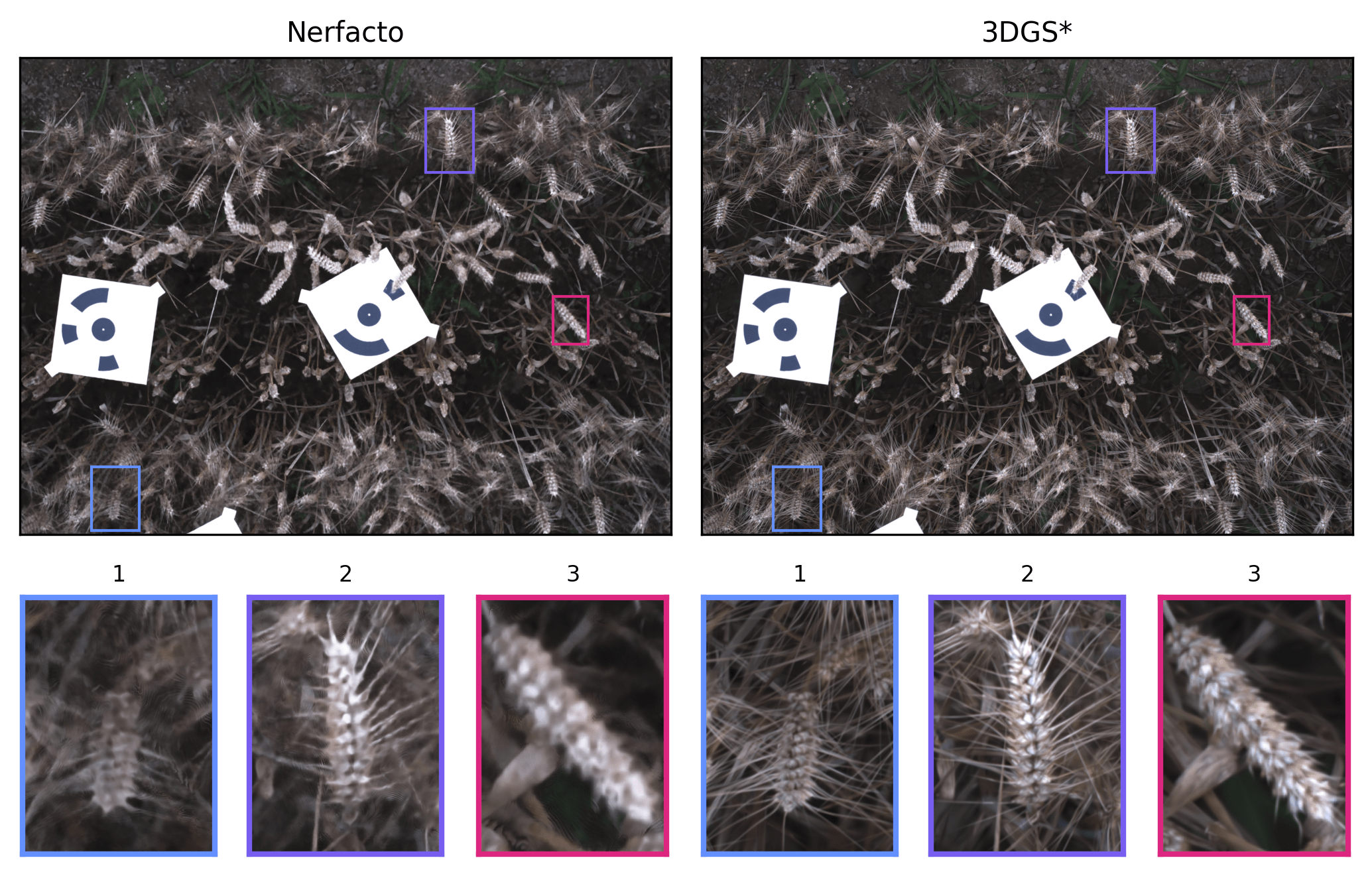}
    \caption{Comparison of Nerfacto and 3DGS* (gsplat implementation) renderings from a test view. Matching zoom regions (1-3) below each image highlight structural details.} 
    \label{fig:wheat_comparison}
\vspace*{-4mm}
\end{figure}

We report Peak Signal-to-Noise Ratio (PSNR), Structural Similarity Index (SSIM) and Learned Perceptual Image Patch Similarity (LPIPS) in \autoref{tab:nvs}, which are standard metrics for quantitative evaluation of visual reconstruction quality of 3D scenes \cite{Arshad24a}.
3DGS (based on gsplat \cite{Ye2024a_gsplat} implementation) achieves the best results on image quality metrics, followed by Nerfacto, with a considerable margin in SSIM and LPIPS. In \autoref{fig:wheat_comparison} we show qualitative comparison, highlighting the greater level of detail achieved by 3DGS, especially on fine-grained structures such as wheat head awns.
Results for additional baselines can be found in the suppl. material.
We note that other baselines based on the original 3DGS codebase present lower results in pixel-wise metrics, but achieve better results in perceptual metrics like LPIPS.
We believe the issue lies in an image transformation problem causing a translation shift by a few pixels for which we could not find a solution.  
This problem only affects 2D evaluations and does not cause visible quality degradation in the 3D reconstructions. We used reconstructions from the original 3DGS method in the rest of our pipeline for ease of integration with our 3D instance segmentation solution.


\input{tables/nvs}

\subsection{Wheat head segmentation}\label{sec:results-seg}



For obtaining 2D input segmentation masks, we experimented with a combination of a pretrained YOLO \&  SAM; and a zero-shot approach based on Grounded SAM \cite{ren2024grounded}.
The Grounded SAM approach used a ``wheat head'' prompt and incorporated SAHI (Slicing Aided Hyper Inference) \cite{akyon2022sahi}, which is designed to improve detections of multiple small objects.
The results of the predictions by these two methods can be seen in \autoref{fig:2dseg_yolo+sam} and \autoref{fig:2dseg_groundedSAM2} respectively.
More wheat heads are detected with the Grounded SAM approach, however, the segmentations are more noisy.
Favoring reliability, we used the pretrained YOLO \& SAM approach in our automated 3D segmentation solution despite the lower amount of detections, given that this limitation can be effectively alleviated with detections from other views. 

To evaluate 3D wheat head segmentation methods on our data we annotated the bounding boxes for each observable wheat head on a randomly-chosen test view per plot.
The wheat heads instances were then segmented by giving these bounding boxes as input prompts to SAM. We visually verified the quality of the output masks. 

We compared our method with FruitNeRF \cite{Meyer24a}, providing the same input segmentation masks.
Quantitative results of both methods compared to the ground truth are presented in \autoref{tab:2dseg}.
Additionally, qualitative comparisons of our method and FruitNeRF to the ground truth mask of one plot are illustrated in \autoref{fig:2dseg_fruitnerf} and \autoref{fig:2dseg_ours}.
Note that many more wheat heads are detected than in the input masks, highlighting the effective incorporation of multi-view detections into our 3D representation.



\input{tables/2dseg}
\input{large_figures/vis2d}







\subsection{Geometric validation and applications} \label{sec:results-fruit_sizing}

\begin{figure*}[tb]
    \centering
    \includegraphics[width=1\linewidth]{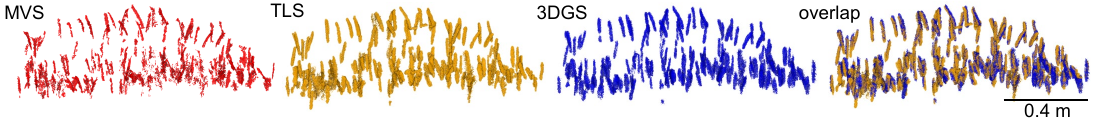}
    \caption{Corresponding wheat head instances of one experimental plot extracted from all three datasets.}
    \label{fig:whs_comparison}
\end{figure*}

We validated the geometric accuracy of our extracted wheat heads by comparing them to aligned TLS data of individual wheat head instances in terms of their 3D morphological traits (length - L, width - W, volume - V), following \autoref{sec:trait_extraction}. We compared the Gaussian centers (i.e. point clouds) of our 3DGS results to TLS and MVS point clouds. However, only the point clouds from our 3DGS-based pipeline were segmented by instance of individual wheat heads. To enable comparison across modalities, we: 1) took single geo-referenced 3DGS wheat heads and filtered out all TLS and MVS data that were $>$15~mm away from the closest 3DGS point; 
2) assigned oriented bounding boxes to each 3DGS wheat head instance, applying a buffer of 10~mm (accounting for the alignment uncertainty), and extracted the matching wheat head instances from TLS and MVS data (\autoref{fig:whs_comparison}).

We compared 3DGS- and MVS-based estimates to TLS on a per-instance and a per-row-average (genotype) basis by linear regression after outlier removal. The per-row-average analysis investigates the potential for phenotyping applications. The results are summarized in \autoref{tab:fruit_sizing}. The observed significant, but moderate correlations ($\rho$) indicate general agreement between the datasets, however, with a high per-instance noise (see e.g. mean absolute percentage error (MAPE) or mean absolute error (MAE)). Computing per-row averages notably decreases MAPE in all cases, increases $\rho$ for L and W, but not for V. These results indicate that: 1) 3DGS and MVS perform comparably well; 2) noise is too high for confident per-instance phenotyping; 3) averaging reduces noise levels to the point that phenotypic data can be used to distinguish different genotypes with moderate to high confidence; 4) the extracted L and W values represent the same real-world physical quantities. Yet, the extracted V values are either too noisy, or image-based and scanning-based estimates capture different aspects of plant structure.

To further test the hypothesis that the extracted traits could be useful for phenotyping applications (i.e. to measure statistically significant differences between genotypes) we conducted a one-way analysis of variance (ANOVA) for each measurement method and present the results in Table \autoref{tab:anova}. The derived F-statistics were significantly ($P$-value $\ll$ 0.01) and notably higher than 1, indicating rejection of the null-hypothesis (no significant difference between genotype means) and strong discriminative power of the derived quantities. For L, TLS provided the largest between-genotypes variability. However, for W and V, 3DGS was notably more discriminative, hinting stronger usability in phenotyping applications than the reference TLS data. 

\input{tables/fruit_sizing}

\input{tables/anova}

%% file: tables/nvs.tex
\begin{table}[htpb]
\centering
\caption{Quantitative comparison for NVS on our test set. We evaluate radiance fields methods (after 30k iterations) based on image quality metrics, average training time, and storage of the trained model. Colors highlight {\color{red} best}, {\color{orange} second-best} and {\color{yellow} third-best} method on each metric. 3DGS*: gsplat implementation of 3DGS.}
\resizebox{1.0\linewidth}{!}{
\begin{tabular}{cccccc}
\toprule
Method & SSIM$\uparrow$ & PSNR$\uparrow$ & LPIPS$\downarrow$ & Time (min) & Storage (GB) \\
\midrule 
Instant-NGP \cite{Mueller22a} & 0.662 & 20.891 & 0.506 & \cellcolor{red!20} 39 & \cellcolor{orange!20} 0.185 \\
Nerfacto \cite{Tancik23a} & \cellcolor{orange!20} 0.769 & \cellcolor{orange!20} 25.387 & \cellcolor{orange!20} 0.384 & \cellcolor{orange!20} 45 & \cellcolor{red!20} 0.164 \\ 
FruitNeRF \cite{Meyer24a} & \cellcolor{yellow!20} 0.752 & \cellcolor{yellow!20} 23.382 & \cellcolor{yellow!20} 0.422 & \cellcolor{yellow!20} 47 & \cellcolor{yellow!20} 0.236 \\
3DGS* \cite{Ye2024a_gsplat} & \cellcolor{red!20} 0.843 & \cellcolor{red!20} 25.447 & \cellcolor{red!20} 0.226 & 146 & 0.557 \\
\bottomrule
\end{tabular}}
\vspace*{-4mm}
\label{tab:nvs}
\end{table}

%% file: tables/2dseg.tex
\renewcommand{\thefootnote}{\fnsymbol{footnote}}

\begin{table}[t]
\centering
\caption{Quantitative results against ground truth 2D segmentation masks. Best results per metric are highlighted in {\color{red} red}.}
\resizebox{1.0\linewidth}{!}{
\begin{tabular}{ccccccc}
\toprule
Method & IoU (\%) & Precision (\%) & Recall (\%) & F1 & MSE & SSIM \\
\midrule 
FruitNeRF \cite{Meyer24a} & 0.34 & \cellcolor{red!20} 0.95 & 0.35 & 0.50 & \cellcolor{red!20} 0.05 & 0.70 \\
Ours & \cellcolor{red!20} 0.50 & 0.81 & \cellcolor{red!20} 0.57 & \cellcolor{red!20} 0.67 & 0.06 & \cellcolor{red!20} 0.90 \\
\bottomrule
\end{tabular}}
\label{tab:2dseg}
\vspace*{-4mm}
\end{table}

%% file: large_figures/vis2d.tex
\begin{figure*}[ht]
\centering
\begin{subfigure}[b]{0.197\linewidth}
    \centering
    \includegraphics[width=1.0\linewidth]{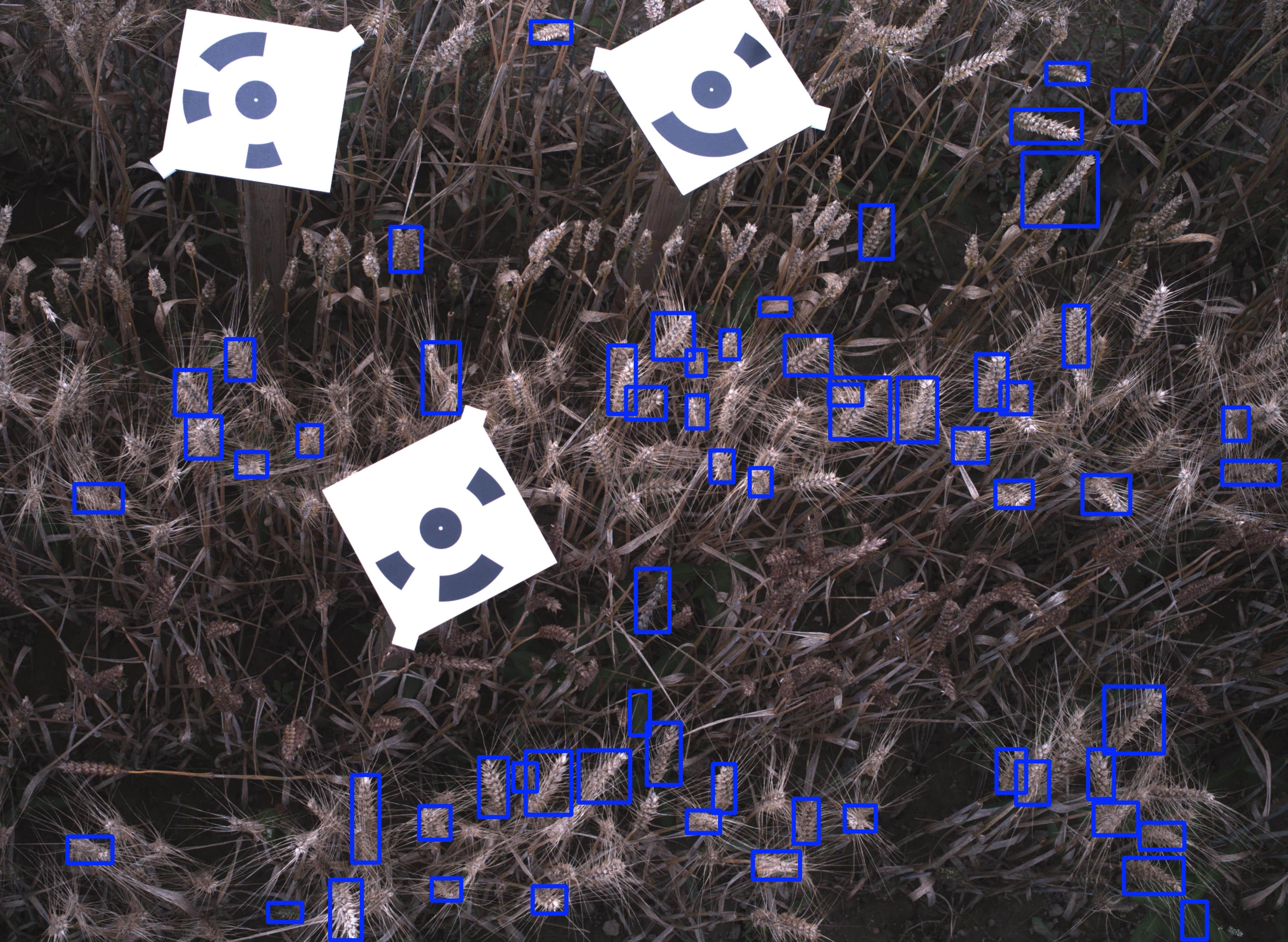}
    \caption{Pretrained YOLO}
    \label{fig:2dseg_yolo}
\end{subfigure}%
\begin{subfigure}[b]{0.197\linewidth}
    \centering
    \includegraphics[width=1.0\linewidth]{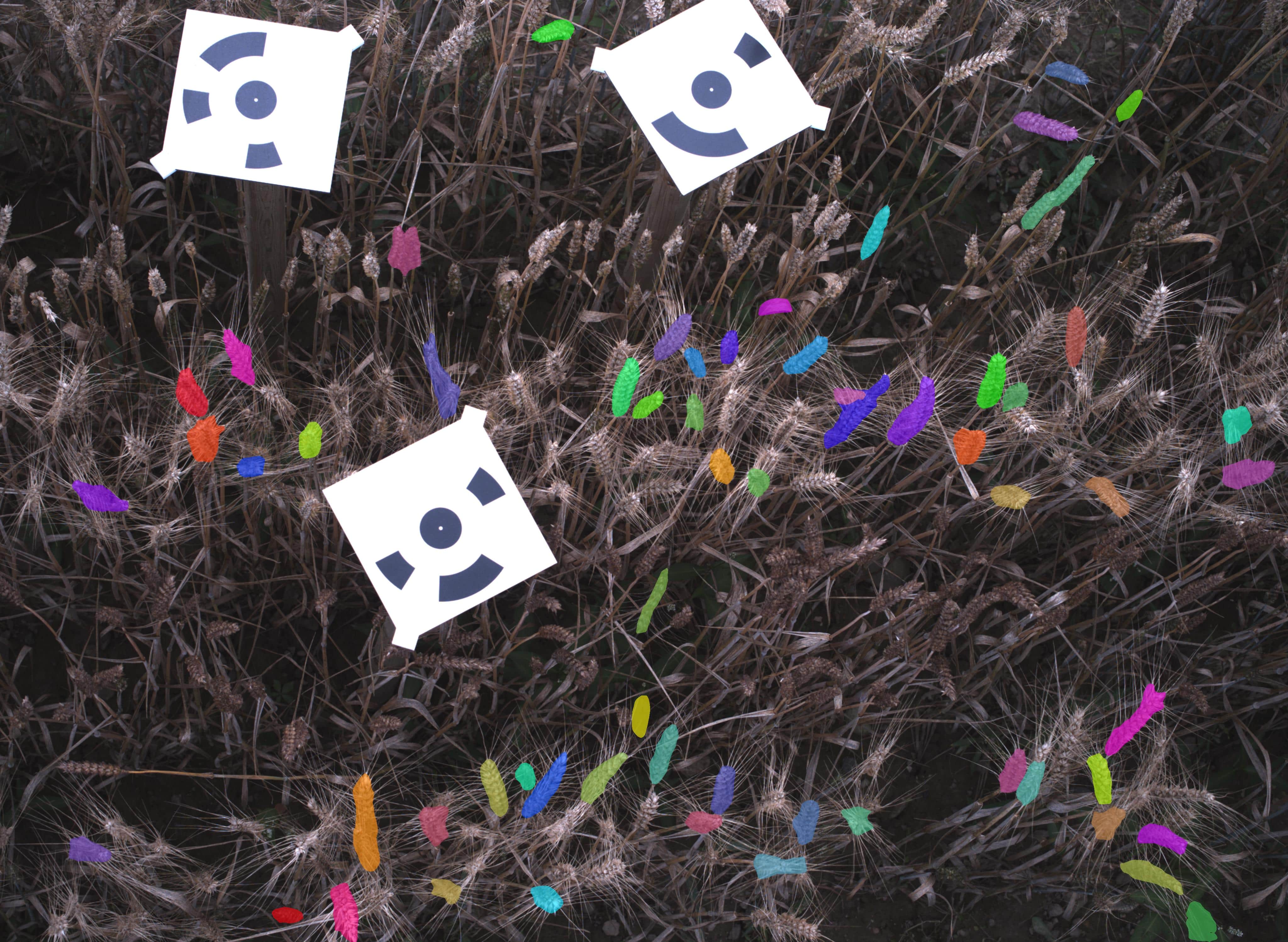}
    \caption{Pretrained YOLO + SAM}
    \label{fig:2dseg_yolo+sam}
\end{subfigure}%
\begin{subfigure}[b]{0.197\linewidth}
    \centering
    \includegraphics[width=1.0\linewidth]{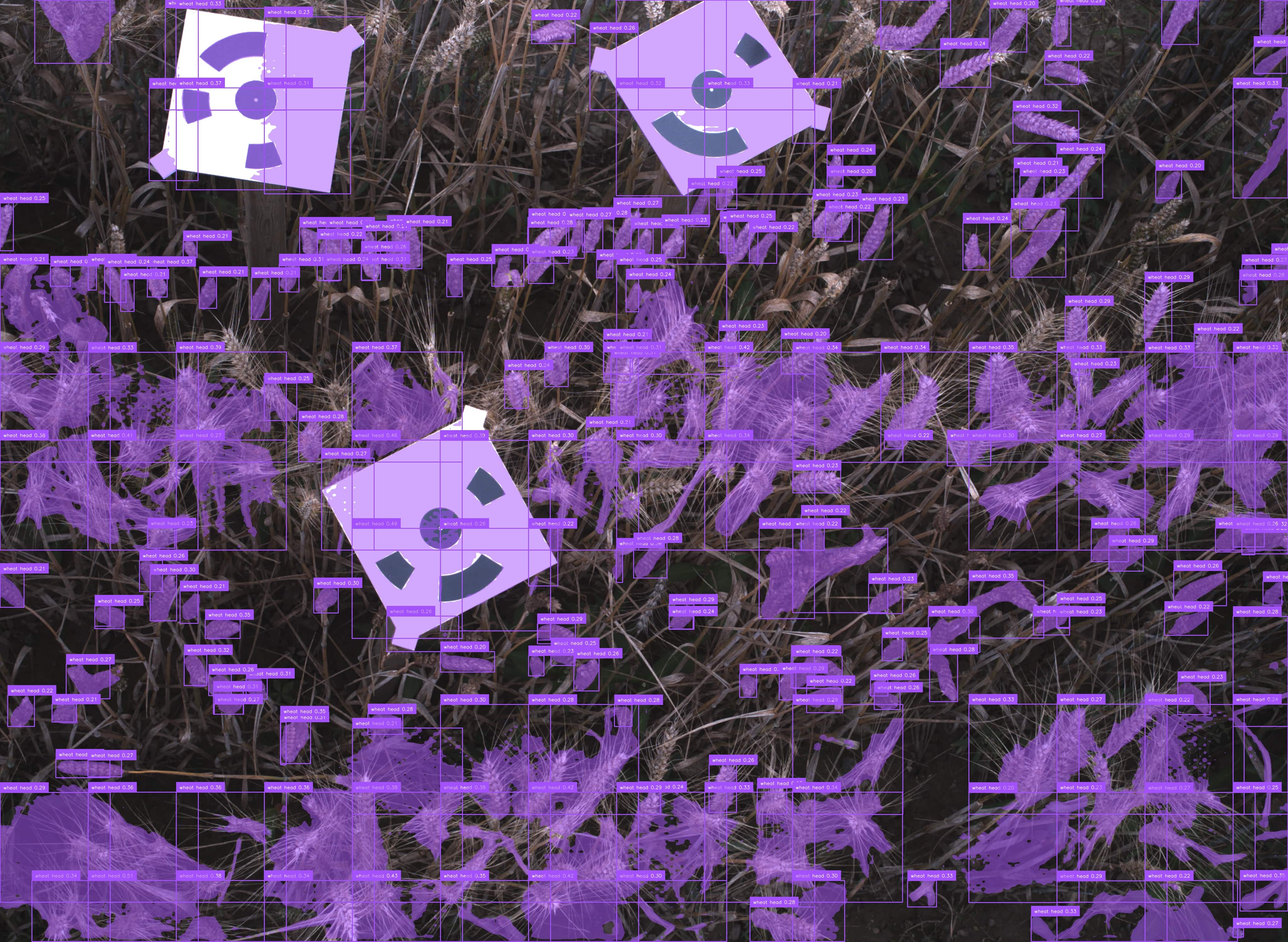}
    \caption{Grounding DINO + SAHI}
    \label{fig:2dseg_groundedSAM2}
\end{subfigure}%
\hspace{0.005mm}
\begin{tikzpicture}
    \draw[dashed, gray, thick] (0,-1.9) -- (0,1.2);
\end{tikzpicture} 
\begin{subfigure}[b]{0.197\linewidth}
    \centering
    \includegraphics[width=1.0\linewidth]{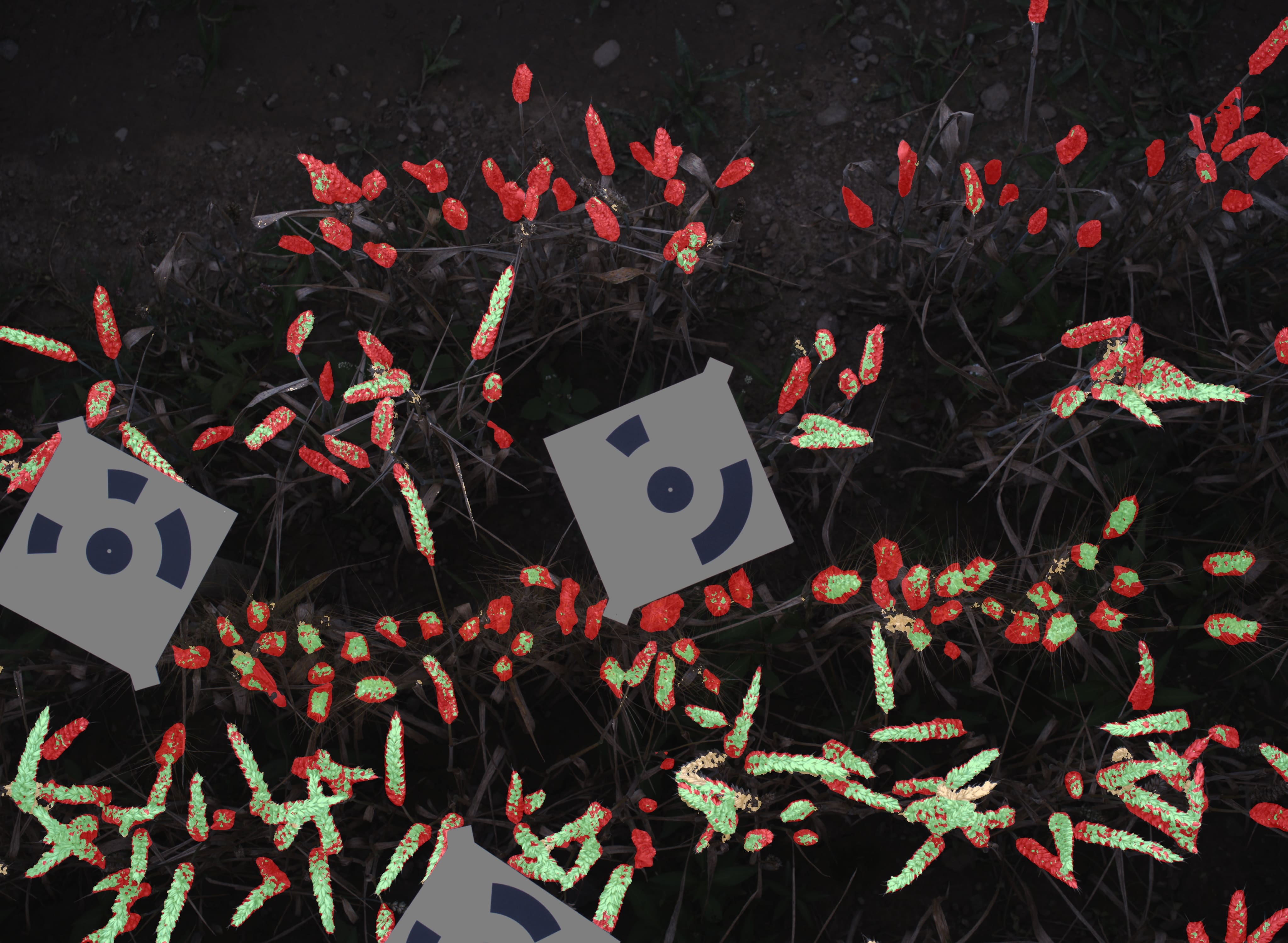}
    \caption{GT vs. FruitNeRF}
    \label{fig:2dseg_fruitnerf}
\end{subfigure}%
\begin{subfigure}[b]{0.197\linewidth}
    \centering
    \includegraphics[width=1.0\linewidth]{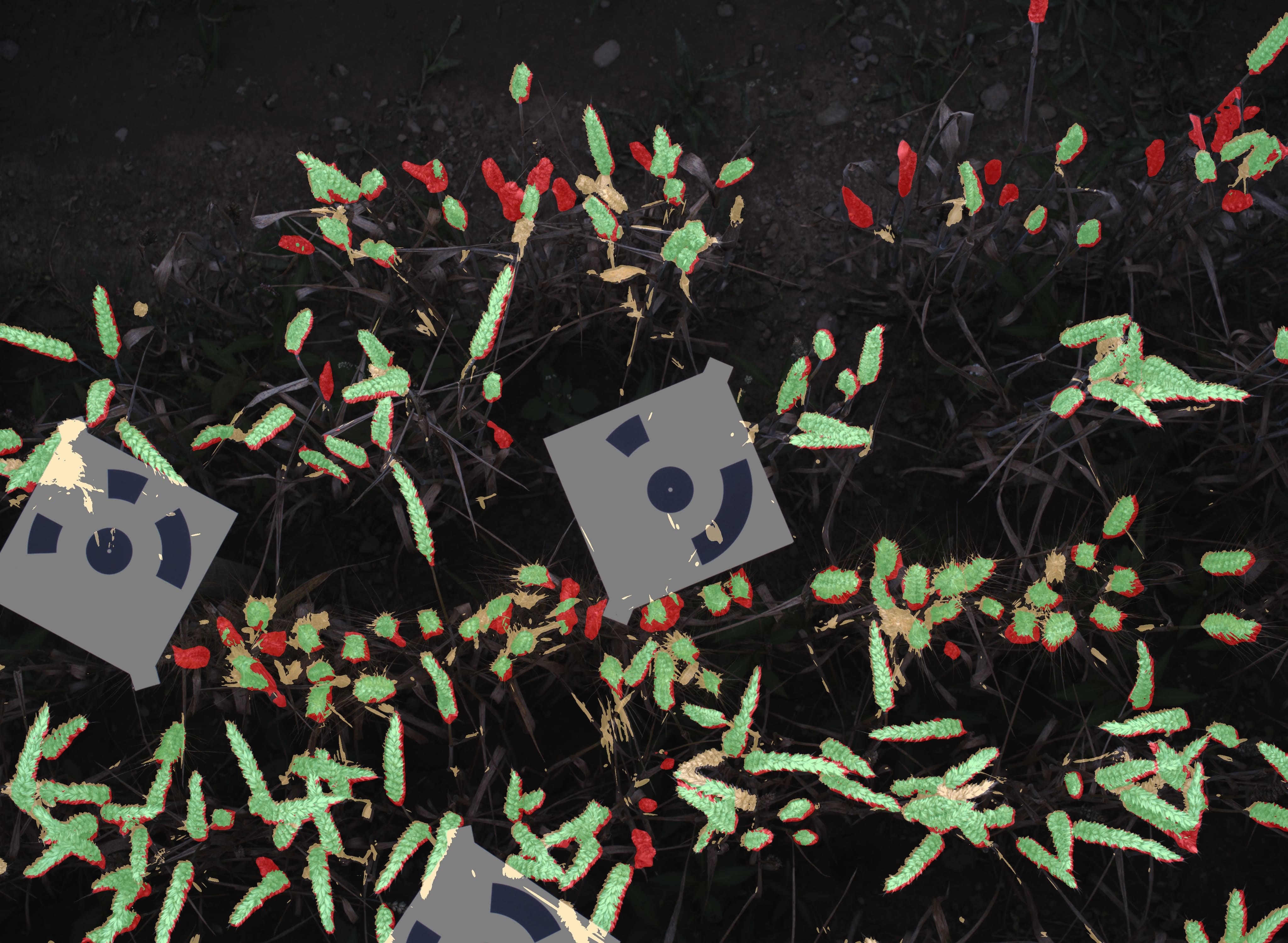}
    \caption{GT vs. Ours}
    \label{fig:2dseg_ours}
\end{subfigure}
\caption{\textbf{Left: visualization of 2D detection and segmentation} of wheat heads on a train image from (a) pre-trained YOLOv5, (b) Segment Anything with detected bounding boxes as prompt, and (c) state-of-the-art GroundedSAM2 version which combines Grounding DINO 1.5 with SAHI (Slicing Aided Hyper Inference), with ``wheat head'' as text prompt. \textbf{Right: qualitative evaluation of novel view mask rendering} from the 3D segmentation obtained by our pipeline (using (b)). (d) and (e) compare the projection of 3D instance segmentation onto 2D in a novel view with human-labeled wheat head segmentation. \textcolor{mygreen}{Green} represents the overlap between rendered masks and ground truth (GT), i.e. correct segmentation of wheat head; {\color{orange}orange} indicates false positive segmentation; and {\color{red}red} represents wheat heads not identified in the 3D instance segmentation, resulting in their absence in the projected 2D masks.}
\label{fig:2dseg_quality}
\end{figure*}

%% file: tables/fruit_sizing.tex
\begin{table}[t]
\centering
\caption{Per-instance and per-row-average agreement: TLS (reference) vs. 3DGS and MVS. We report correlation ($\rho$), mean absolute error (MAE), and mean absolute percentage error (MAPE) for length (L), width (W), volume (V). MAE units are in cm for L and W, and cm$^3$ for V. P-value $\ll0.01$ in each per-instance case, $\le0.05$ in each per-row-average case, except 3DGS-V. Best results per trait and metric are highlighted in {\color{red} red}.}
\small
\resizebox{0.7\linewidth}{!}{

\begin{tabular}{@{}llcccccc@{}}
    \toprule
    & & \multicolumn{3}{c}{per-instance} & \multicolumn{3}{c}{per-row-average} \\
    \cmidrule(lr){3-5}\cmidrule(lr){6-8}
    & & L & W & V & L & W & V \\
    \midrule
    $\rho$ & MVS & \cellcolor{red!20} 0.51 & \cellcolor{red!20} 0.35 &  \cellcolor{red!20} 0.40 & \cellcolor{red!20} 0.74 & \cellcolor{red!20} 0.53 & \cellcolor{red!20} 0.32 \\
     & 3DGS  & \cellcolor{red!20} 0.51 & 0.27 &  0.32 & 0.69 & 0.43 & 0.05 \\
    \midrule
     MAE  & MVS & 1.51 & 0.35 & 12.57 & \cellcolor{red!20} 0.58 & 0.19 & 9.64 \\
     & 3DGS  & \cellcolor{red!20} 1.48 & \cellcolor{red!20} 0.25 & \cellcolor{red!20} 10.72 & 0.79 & \cellcolor{red!20} 0.13 & \cellcolor{red!20} 6.12 \\
     \midrule
    MAPE & MVS & 16.0 & 26.0 & 47.2 & \cellcolor{red!20} 5.9 & 15.0 & 39.9 \\
    & 3DGS  & \cellcolor{red!20} 15.1 & \cellcolor{red!20} 18.3 & \cellcolor{red!20} 40.2 & 8.1 & \cellcolor{red!20} 9.9 & \cellcolor{red!20} 24.4 \\
    \bottomrule
\end{tabular}
}
\label{tab:fruit_sizing}
\end{table} 

%% file: tables/anova.tex
\begin{table}[htbp]
\centering
\caption{One-way ANOVA F-statistics for length (L), width (W), and volume (V) based on 2389 samples of 42 populations (P-value $\ll0.01$ in each case). Best results per trait are highlighted in {\color{red} red}.}
\small
\resizebox{0.75\linewidth}{!}{
    \begin{tabular}{@{}ccc ccc ccc@{}}
        \toprule
        \multicolumn{3}{c}{L} & \multicolumn{3}{c}{W} & \multicolumn{3}{c}{V} \\
        \cmidrule(lr){1-3}\cmidrule(lr){4-6}\cmidrule(lr){7-9}
        TLS & 3DGS & MVS & TLS & 3DGS & MVS & TLS & 3DGS & MVS \\
        \midrule
        \cellcolor{red!20} 15.2 & 11.2 & 10.1 & 5.2 & \cellcolor{red!20} 35.0 & 8.5 & 6.9 & \cellcolor{red!20} 10.8 & 7.1 \\
        \bottomrule
    \end{tabular}}
\vspace*{-4mm}
\label{tab:anova}
\end{table}

%% file: sec/6_discussion.tex
\section{Discussion} \label{sec:discussion}

We showed that the combination of 3DGS with our multi-view instance segmentation pipeline effectively captures the complex geometry of wheat canopies allowing to extract, count and measure hundreds of individual wheat heads in 3D. Unlike previous RGB image-based approaches for 3D reconstruction that typically require dense spatial coverage of viewpoints, our method achieved high-quality reconstructions with only 30 views per plot and with a limited viewpoint distribution (\autoref{fig:title}). This is particularly important for practical field applications, where capturing dense and diverse viewpoints may be infeasible or time-consuming. Comparatively, NeRF-based methods like FruitNeRF \cite{Meyer24a} tackled simpler scenes and tasks (fruit counting in horticulture) using several hundreds of images with good spatial coverage. Meanwhile, \cite{GeneMola23_looking} proposed an MVS-based approach to measure fruits from 2D images with precise amodal segmentation masks and depth maps, but cannot perform volume estimation. 

Beyond good 3D reconstruction and segmentation, we show that our method allows for in-field morphological trait extraction with a sufficient precision for distinguishing between different genotypes, facilitating phenotyping applications in a breeding context. So far, similar achievements have been demonstrated only using expensive high-end laser scanning instruments ~\cite{liu2023extraction,wang2022unsupervised}, attaining comparable results with lower uncertainty (MAPE per-instance: L of 4-5\% and W of 12-32\%; MAPE per-genotype-average: L of 15\% and W of 24\%).

\begin{figure}[htbp]
    \centering
    \includegraphics[width=0.8\linewidth]{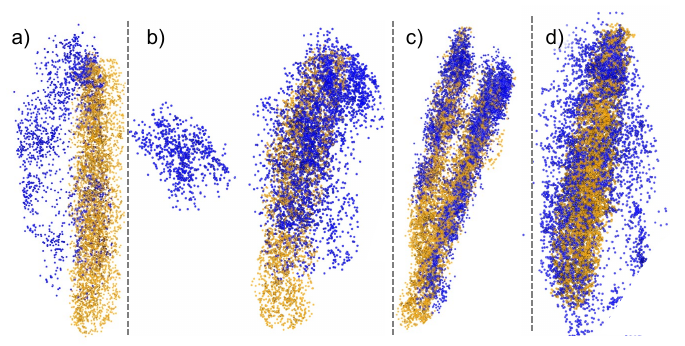}
    \caption{Reoccurring failure cases causing disagreement between the reference TLS (orange) and 3DGS (blue): a) incorrect reconstruction at lower canopy levels and scene edges; b) and c) incorrect instance segmentation; d) ambigous geometry reconstruction due to strongly expressed awns.}
    \vspace*{-5mm}
    \label{fig:failure_cases}
\end{figure}

Despite reasonably promising results, we observed some disagreements between 3DGS and the reference TLS. There are several common failure cases (\autoref{fig:failure_cases}): a) splats of wheat heads at the lower canopy levels and on the scene edges (limited number of views) can diverge from the real wheat head surface and can fail to reconstruct the bottom part of the wheat head; b) occasional errors in instance segmentation lead to multiple splat clusters being related to a single instance---sometimes capturing wheat head-unrelated foliage, or c) merging multiple wheat heads together; d) splats partially capturing structure of strongly expressed awns (spikes) for some wheat varieties, but insufficiently well for their full reconstruction. The latter phenomena is a likely cause for the observed strong disparity between 3DGS- and TLS-based volume estimates, as the TLS is inherently unable to capture such fine structural details due to finite laser beam footprint size (\autoref{tab:fruit_sizing}). Future work directions to address these issues may include introducing prior knowledge about wheat heads shape in a similar way to \cite{Meyer24a,Magistri24_shape-completion}, and ensuring robustness to environmental disturbances such as wind.

\vspace*{-2mm}

%% file: sec/7_conclusion.tex
\section{Conclusion}
In this work, we address 3D reconstruction of wheat canopies and instance segmentation of wheat heads from multi-view images, using 3D Gaussian Splatting (3DGS), a pretrained wheat head detector, and the Segment Anything Model (SAM). We handle instance segmentation iteratively, by annotating Gaussians using maximal information available from inconsistent binary segmentation masks across views. Our approach demonstrates the effectiveness of 3DGS for the 3D reconstruction of wheat canopies and instance segmentation of wheat heads in field conditions. 
Comparisons against state-of-the-art NeRF-based methods for this task highlight superior reconstruction quality and segmentation performance of our approach. 
Furthermore, evaluations against terrestrial laser scan data demonstrate that our method achieves sufficient accuracy for high-throughput field phenotyping of wheat head morphological traits, including length, width, and volume.

%% file: sec/X_suppl.tex
\clearpage
\setcounter{page}{1}
\maketitlesupplementary

\section{Dataset setup details}
\begin{figure}[h]
    \centering
    \includegraphics[width=\linewidth]{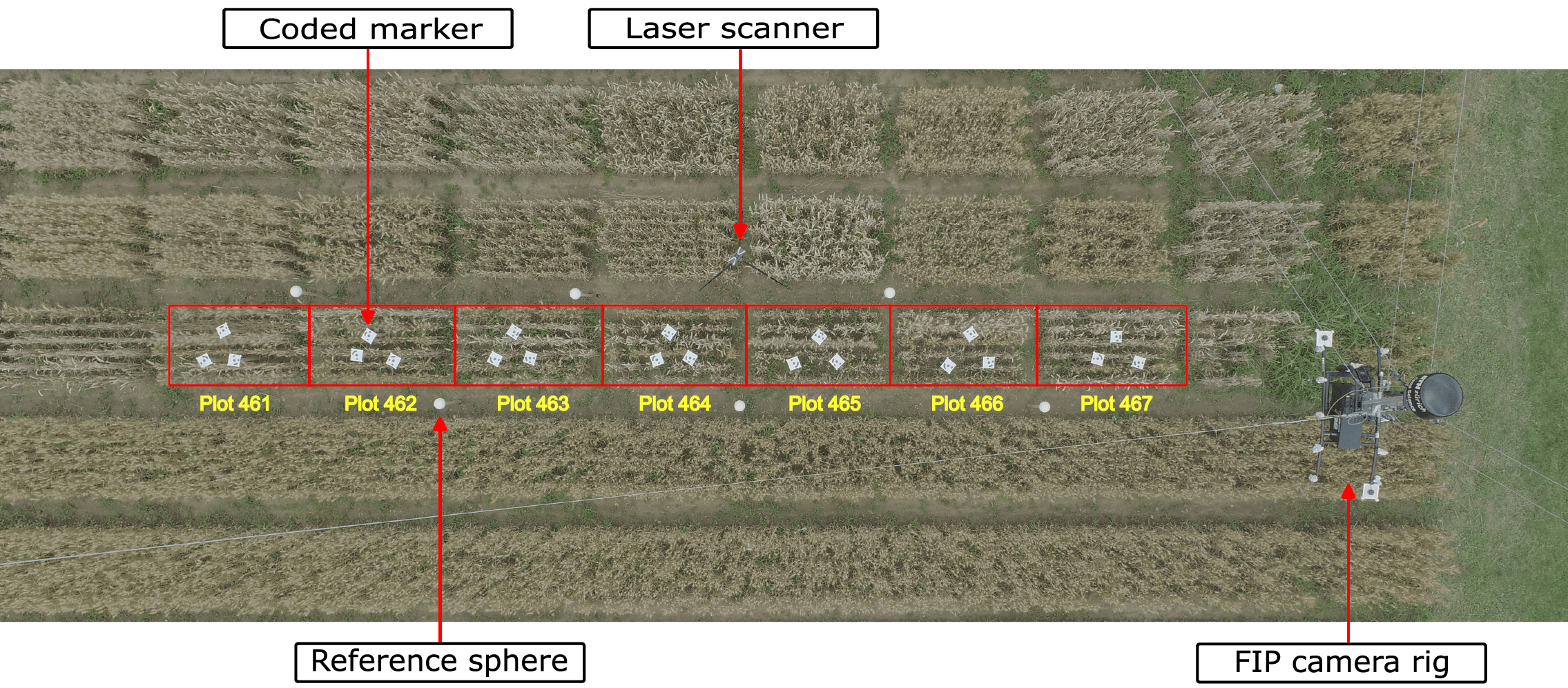}
    \caption{Overview of our data collection setup.}
    \label{fig:data_collection}
\end{figure}

\begin{figure}[htbp]
    \centering
    \includegraphics[width=\linewidth]{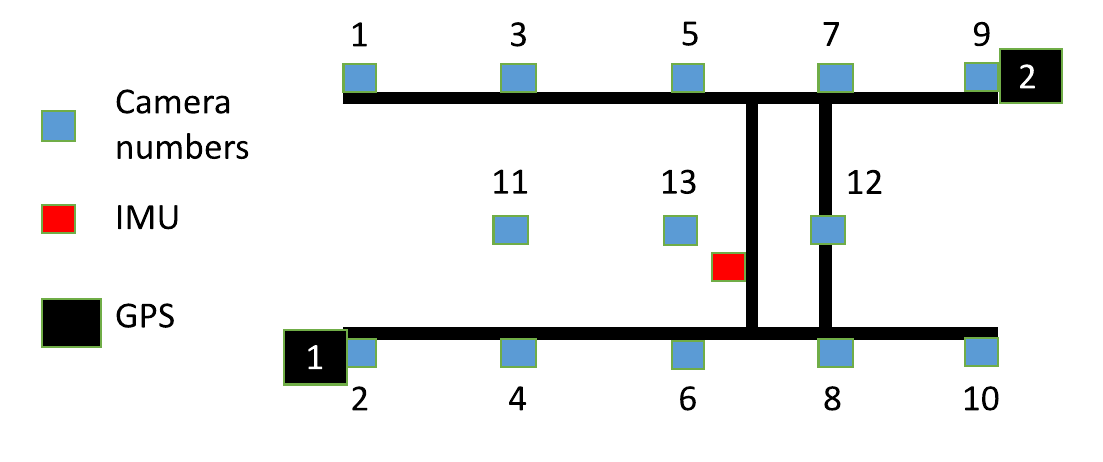}
    \caption{Schematic top view of the FIP multi-camera rig system.}
    \label{fig:rig_diagram}
\end{figure}

\begin{figure}[htbp]
    \centering
    \includegraphics[width=\linewidth]{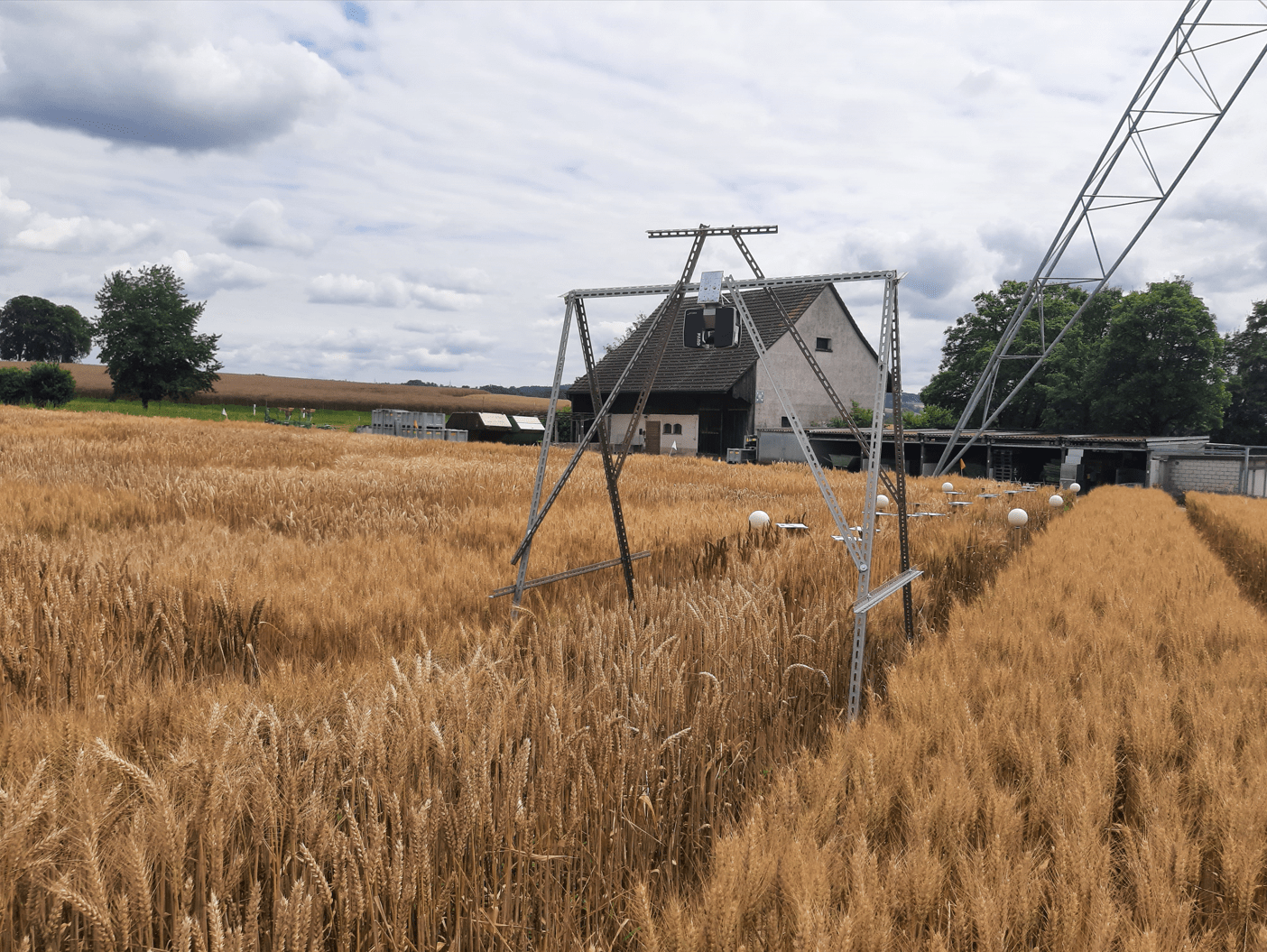}
    \caption{Our custom mount for upside-down laser scans.}
    \label{fig:scanner_mount}
\end{figure}

We present an overview of our data collection setup on seven wheat plots in \autoref{fig:data_collection}. Each wheat plot contained six rows of different wheat varieties. Image acquisition was performed using the Field Phenotyping Platform (FIP) of ETH Zürich \cite{Kirchgessner2017}. The platform consists of a multi-view camera rig mounted on a SpiderCam (Spidercam robotics GmbH, Feistritz, Austria) cable system and is equipped with 13 cameras (\autoref{fig:rig_diagram}). We used only 12 cameras (DFK 38UX304, 12 MP, The Imaging Source, Bremen, Germany) for their identical lens specifications (V3522-MPZ, 35 mm, The Imaging Source, Bremen, Germany) and field of view (FOV). For each of the seven plots, we captured three sets of 12 images, with an approximately 25 cm collinear shift in rig position between the sets, resulting in 36 views per plot for 3D reconstruction. Three coded ring markers were placed on each plot to aid Structure from Motion (SfM), set the scale, and enable alignment of reference laser scans. Marker coordinates were measured with a Trimble R10 GNSS device in RTK mode (1-2 cm positioning accuracy). SfM was performed in Agisoft Metashape (St. Petersburg, Russia) using all 36 images to obtain camera calibrations and a sparse point cloud per plot. This provided a basic input for our proposed workflow. In addition, the MVS pipeline was performed to obtain dense point clouds for comparing the proposed workflow against the traditional 3D photogrammetry reconstruction. Finally, \autoref{fig:scanner_mount} shows the custom laser scanner mount used in this study in addition to the tripod shown in \autoref{fig:data_collection}.

\section{Additional NVS baselines}
While gsplat \cite{Ye2024a_gsplat} implementation of 3DGS outperforms other radiance field methods in terms of NVS, we also experimented with the original 3DGS implementation by Inria \cite{Kerbl23a}, as well as 2D Gaussian Splatting (2DGS) \cite{Huang24a} and SuGaR \cite{Guedon23a} (\autoref{tab:nvs2}). Notably, the latter two methods provide advantages for 3D mesh extraction, which may be desirable in certain workflows. As described in \autoref{sec:results-nvs}, we observed a pixel misalignment between the rendered evaluation views and ground truth views when using the original implementation of 3DGS and its variants, 2DGS and SuGaR. Such positional shifts significantly degrade pixel-wise image quality metrics, such as SSIM and PSNR, causing it to perform worse than NeRF-based methods, despite achieving higher perceptual image quality metrics like LPIPS.
Regarding NeRF-based models, although FruitNeRF-big \cite{Meyer24a} has greater layer depth, hidden dimension, and overall capacity for modeling density and appearance, its performance degrades compared to the smaller version. We suspect the reason is that our limited amount of training views (30) compared to the original dataset the model was developed on leads to severe overfitting \cite{Meyer24a}.

\begin{table}[htpb]
\centering
\caption{\textbf{Quantitative comparison for Novel View Synthesis} on our dataset. We evaluate neural rendering methods based on image quality metrics, average training time, and stored model size. 3DGS*: gsplat implementation of 3DGS. Note that pixel-wise metrics (SSIM, PSNR) for 3DGS \cite{Kerbl23a} and its variants, 2DGS \cite{Huang24a} and SuGaR \cite{Guedon23a}, are negatively affected by pixel misalignment between rendered and ground truth views due to a bug in data transformation, which does not severely affect patch-based metrics (LPIPS). The two types of SuGaR (coarse and refined) correspond to the sets of 3D Gaussians extracted at different stages of SuGaR’s optimization for surface alignment.}
\resizebox{1.0\linewidth}{!}{
\begin{tabular}{ccccccc}
\toprule
Type & Method & SSIM$\uparrow$ & PSNR$\uparrow$ & LPIPS$\downarrow$ & Time (min) & Storage (GB) \\
\midrule 
\multirow{3}{*}{\shortstack{NeRF- \\ based}}
& Instant-NGP \cite{Mueller22a} & 0.662 & 20.891 & 0.506 & 39 & 0.185 \\
& Nerfacto \cite{Tancik23a} & 0.769 & 25.387 & 0.384 & 45 & 0.164 \\ 
& FruitNeRF \cite{Meyer24a} & 0.752 & 23.382 & 0.422 & 47 & 0.236 \\
& FruitNeRF {\footnotesize{big}} & 0.500 & 15.663 & 0.666 & 440 & 0.792\\
\midrule
\multirow{7}{*}{\shortstack{Gaussian- \\ based}} 
& 3DGS* \cite{Ye2024a_gsplat} &  0.843 &  25.447 &  0.226 & 146  & 0.557 \\
& 3DGS {\footnotesize{7k iters}} \cite{Kerbl23a} &  0.651 & 20.549 & 0.333 & 31 & 0.996\\
& 3DGS {\footnotesize{15k iters}} &  0.639 & 20.416 & 0.323 & 74 & 1.286\\
& 2DGS \cite{Huang24a} & 0.560 & 20.593 & 0.241 & 72 & - \\
& SuGaR {\footnotesize{coarse}} \cite{Guedon23a} & 0.569 & 20.716 & 0.278 & 40 & 0.102\\
& SuGaR {\footnotesize{refined}} & 0.549 & 20.520 & 0.290 & 84 & 0.488\\
\bottomrule
\end{tabular}}

\label{tab:nvs2}
\end{table}

\section{Additional qualitative results}

\begin{figure}[htpb]
    \centering
    \includegraphics[width=\columnwidth]{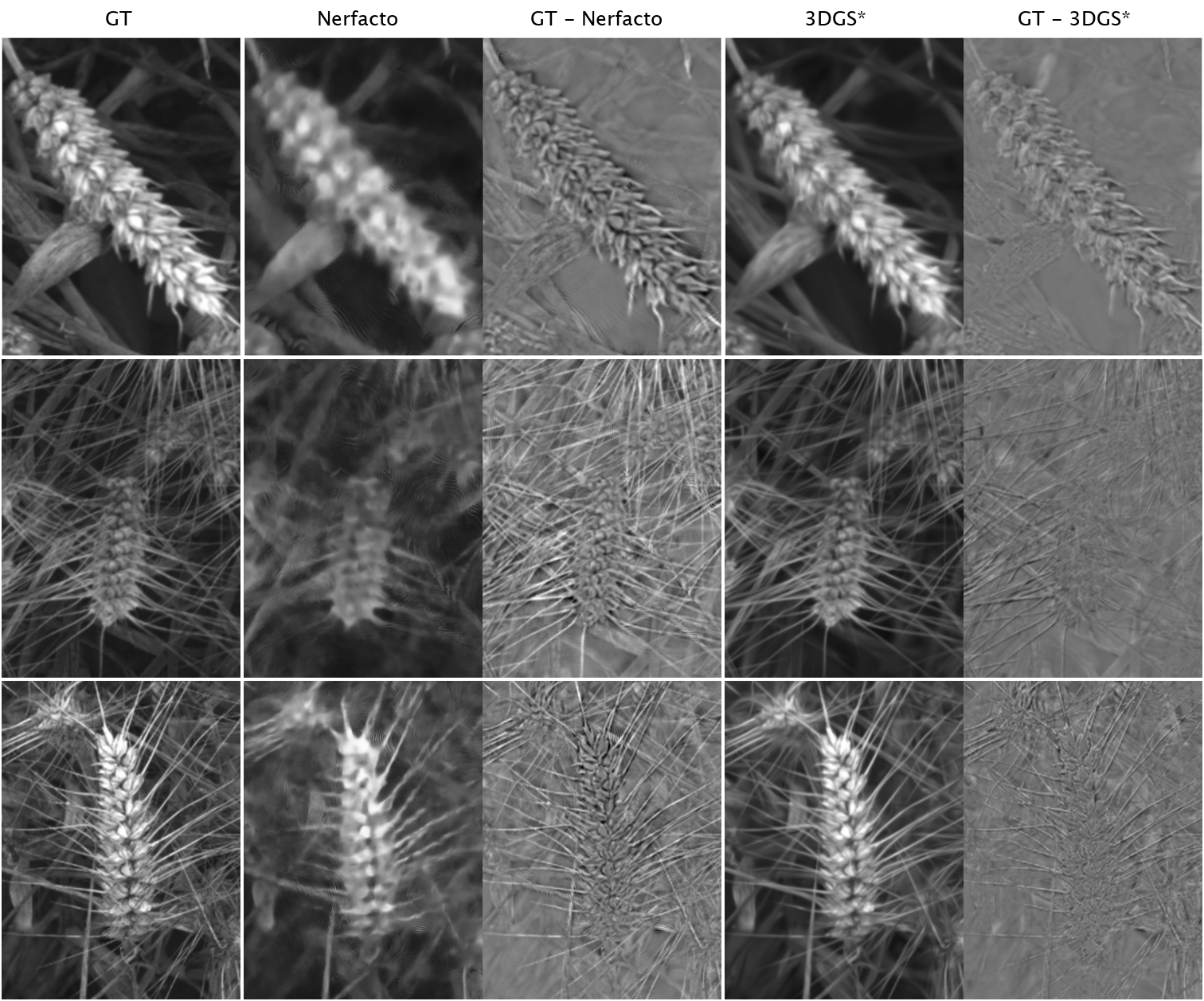}
    \caption{Comparison of wheat head renderings (same ones as in \autoref{fig:wheat_comparison}) to the ground truth image. From left to right: ground truth (GT), Nerfacto, GT-Nerfacto difference, 3DGS* (gsplat implementation), and GT-3DGS* difference. All difference maps are shown in identical grayscale range.} 
    \label{fig:grayvalue_rendering_differences}
    \vspace*{-4mm}
\end{figure}

We provide additional qualitative results in \autoref{fig:grayvalue_rendering_differences}, which demonstrate that 3DGS* produces renderings with fewer deviations from the ground truth image compared to Nerfacto, as evidenced by the reduced structural details visible in the difference maps.

\section{Additional laser scan comparisons}
We repeated the comparison of the 3DGS, TLS and MVS-based wheat head length (L), width (W) and volume (V) estimates, per-instance and per-row average (genotype), after removing obvious 3D reconstruction failure cases as discussed in \autoref{sec:discussion}. 

Failure cases were automatically detected as out-of-the-distribution samples of 2D L, W, and V values distributions; where the first and second dimensions of the respective 2D distributions were defined as TLS-based and 3DGS-based trait estimate values. The expected (failure-case-free) theoretical data distributions were determined by robustly fitting 2D Gaussians (by Minimum Covariance Determinant - MCD estimator) and detecting and removing all points that were outside the confidence interval (comparing squared mahalanobis distance with threshold value drawn from the Chi-squared distribution, 95th percentile, 2 degrees of freedom).

The updated results with mainly improved metrics are presented in \autoref{tab:fruit_sizing_supplement}. Eliminating the most prominent of these failure cases leads to notable increases in similarity between 3DGS-based and TLS-based estimates. 
MAE decreases from 1.48 to 0.73 cm, 0.25 to 0.21 cm, and 10.72 to 7.25 cm$^3$ for the per-instance comparison case for L, W and V respectively; and changes from 0.79 to 0.52 cm, 0.13 to 0.11 cm, and 6.12 to 4.48 cm$^3$ for the per-row-average case (on average MAE decreases 30$\%$).

\input{tables/fruit_sizing_supplement}

%% file: tables/fruit_sizing_supplement.tex
\begin{table}[htpb]
\centering
\caption{Per-instance and per-row-average agreement after filtering out the failure cases: TLS (reference) vs. 3DGS and MVS. We report correlation ($\rho$), mean absolute error (MAE), and mean absolute percentage error (MAPE) for length (L), width (W), volume (V). MAE units are in cm for L and W, and cm$^3$ for V. P-value $\ll0.01$ in each per-instance case, $\le0.05$ in each per-row-average case, except MVS-V. Best results per trait and metric are highlighted in {\color{red} red}.}
\small
\resizebox{0.7\linewidth}{!}{
\begin{tabular}{@{}llcccccc@{}}
    \toprule
    & & \multicolumn{3}{c}{per-instance} & \multicolumn{3}{c}{per-row-average} \\
    \cmidrule(lr){3-5}\cmidrule(lr){6-8}
    & & L & W & V & L & W & V \\
    \midrule
    $\rho$  & MVS  & 0.55 & \cellcolor{red!20} 0.35 & 0.36 & \cellcolor{red!20} 0.75 & 0.55 & 0.31 \\  
            & 3DGS & \cellcolor{red!20} 0.78 & 0.33 & \cellcolor{red!20} 0.39 & 0.73 & \cellcolor{red!20} 0.58 & \cellcolor{red!20} 0.39 \\
    \midrule
     MAE    & MVS  & 1.09 & 0.31 & 10.00 & \cellcolor{red!20} 0.51 & 0.18 & 8.42 \\
            & 3DGS & \cellcolor{red!20} 0.73 & \cellcolor{red!20} 0.21 & \cellcolor{red!20} 7.25 & 0.52 & \cellcolor{red!20} 0.11 & \cellcolor{red!20} 4.48 \\
     \midrule
    MAPE    & MVS  & 12.3 & 24.1 & 43.9 & \cellcolor{red!20} 5.5 & 14.38 & 38.92 \\
            & 3DGS & \cellcolor{red!20} 8.2 & \cellcolor{red!20} 16.7 & \cellcolor{red!20} 32.15 & 5.6 & \cellcolor{red!20} 8.88 & \cellcolor{red!20} 20.51 \\
    \bottomrule
\end{tabular}
}

\label{tab:fruit_sizing_supplement}
\end{table}